\documentclass[twoside,11pt]{article}

\usepackage[nohyperref]{jmlr2e}
\usepackage{siunitx}
\usepackage{graphicx}
\usepackage{caption}
\usepackage{subcaption}
\usepackage{multicol}
\usepackage{placeins}

\usepackage[para]{footmisc} %

\usepackage[usenames,dvipsnames]{xcolor}
\definecolor{shadecolor}{gray}{0.9}

\usepackage{natbib}
\usepackage[colorlinks,linktoc=all]{hyperref}
\usepackage[all]{hypcap}
\hypersetup{citecolor=MidnightBlue}
\hypersetup{linkcolor=MidnightBlue}
\hypersetup{urlcolor=MidnightBlue}

\usepackage[nameinlink]{cleveref}
\creflabelformat{equation}{#2\textup{#1}#3}  %

\ShortHeadings{Uncertainty Baselines}{}
\firstpageno{1}

\begin{document}

\title{Uncertainty Baselines:\\Benchmarks for Uncertainty \& Robustness in Deep Learning}

\author{\name Zachary Nado \email znado@google.com\vspace{-0.4em}\thanks{Google Research, Brain Team}
\AND
\name Neil Band \email neil.band@cs.ox.ac.uk\vspace{-0.4em}\thanks{OATML, Department of Computer Science, University of Oxford}
\AND
\name Mark Collier \email markcollier@google.com\vspace{-0.4em}\footnotemark[1]
\AND
\name Josip Djolonga \email josipd@google.com\vspace{-0.4em}\footnotemark[1]
\AND
\name Michael W. Dusenberry \email dusenberrymw@google.com\vspace{-0.4em}\footnotemark[1]
\AND
\name Sebastian Farquhar \email sebastian.farquhar@cs.ox.ac.uk\vspace{-0.4em}\footnotemark[2]
\AND
\name Qixuan Feng \email qixuan.feng1@gmail.com\vspace{-0.4em}\footnotemark[2]
\AND
\name Angelos Filos \email angelos.filos@cs.ox.ac.uk\vspace{-0.4em}\footnotemark[2]
\AND
\name Marton Havasi \email mh740@cam.ac.uk\vspace{-0.4em}\thanks{Department of Engineering University of Cambridge}
\AND
\name Rodolphe Jenatton \email rjenatton@google.com\vspace{-0.4em}\footnotemark[1]
\AND
\name Ghassen Jerfel \email ghassen@google.com\vspace{-0.4em}\footnotemark[1]
\AND
\name Jeremiah Liu \email jereliu@google.com\vspace{-0.4em}\footnotemark[1]\thanks{Department of Biostatistics, Harvard University}
\AND
\name Zelda Mariet \email zmariet@google.com\vspace{-0.4em}\footnotemark[1]
\AND
\name Jeremy Nixon \email jnixon2@gmail.com\vspace{-0.4em}\footnotemark[1]
\AND
\name Shreyas Padhy \email shreyaspadhy@google.com\vspace{-0.4em}\footnotemark[1]
\AND
\name Jie Ren \email jjren@google.com\vspace{-0.4em}\footnotemark[1]
\AND
\name Tim G. J. Rudner \email tim.rudner@cs.ox.ac.uk\vspace{-0.4em}\footnotemark[2]
\AND
\name Faris Sbahi \email farissbahi@google.com\vspace{-0.4em}\footnotemark[1]
\AND
\name Yeming Wen \email ywen@utexas.edu\vspace{-0.4em}\thanks{University of Texas at Austin}
\AND
\name Florian Wenzel \email florianwenzel@google.com\vspace{-0.4em}\footnotemark[1]
\AND
\name Kevin Murphy \email kpmurphy@google.com\vspace{-0.4em}\footnotemark[1]
\AND
\name D. Sculley \email dsculley@google.com\vspace{-0.4em}\footnotemark[1]
\AND
\name Balaji Lakshminarayanan \email balajiln@google.com\vspace{-0.4em}\footnotemark[1]
\AND
\name Jasper Snoek \email jsnoek@google.com\vspace{-0.4em}\footnotemark[1]
\AND
\name Yarin Gal \email yarin@cs.ox.ac.uk\vspace{-0.4em}\footnotemark[2]
\AND
\name Dustin Tran \email trandustin@google.com\vspace{-0.4em}\footnotemark[1]
\vspace{-2ex}
}

\editor{}

\maketitle

\vspace{-5ex}
\begin{abstract}%
High-quality estimates of uncertainty and robustness are crucial for numerous real-world applications, especially for deep learning which underlies many deployed ML systems. The ability to compare techniques for improving these estimates is therefore very important for research and practice alike. Yet, competitive comparisons of methods are often lacking due to a range of reasons, including: compute availability for extensive tuning, incorporation of sufficiently many baselines, and concrete documentation for reproducibility. In this paper we introduce Uncertainty Baselines: high-quality implementations of standard and state-of-the-art deep learning methods on a variety of tasks. As of this writing, the collection spans 19 methods across 9 tasks, each with at least 5 metrics. Each baseline is a self-contained experiment pipeline with easily reusable and extendable components. Our goal is to provide immediate starting points for experimentation with new methods or applications. Additionally we provide  model checkpoints, experiment outputs as Python notebooks, and leaderboards for comparing results.
\url{https://github.com/google/uncertainty-baselines}
\end{abstract}

\section{Introduction}

Baselines on standardized benchmarks are crucial to machine learning research for measuring whether new ideas yield meaningful progress. However, reproducing the results from previous works can be extremely challenging, especially when only reading the paper text \citep{sinha2020neurips,d2020underspecification}. Having access to the code for experiments is more useful, assuming it is well-documented and maintained.
But even this is not enough.
In fact, in retrospective analyses over a collection of works, authors often find that a simpler baseline works best in practice, due to flawed experiment protocols or insufficient tuning
\citep{melis2017state,kurach2019large,bello2021revisiting,nado2021large}%
.

There is a wide spectrum of experiment artifacts made available in papers. A popular approach is a GitHub dump of code used to run experiments, albeit lacking documentation and tests.
At best, papers might provide actively maintained repositories with 
examples, model checkpoints, 
and ample documentation to extend the work. 
A single paper can only go so far however: without community standards, each paper's codebase differs in experimental protocol and code organization, making it difficult to compare across papers within a common benchmark, let alone build jointly on top of multiple papers.

To address these challenges, we created the Uncertainty Baselines library. It provides high-quality implementations of baselines across many uncertainty and out-of-distribution robustness tasks. 
Each baseline is designed to be self-contained (i.e., minimal dependencies) and easily extensible. We provide numerous artifacts in addition to the raw code so that others can adapt any baseline to suit their workflow.

\textbf{Related work.}
OpenAI Baselines \citep{dhariwal2017openai} is work in similar spirit for reinforcement learning. Prior work on uncertainty and robustness benchmarks include \citet{riquelme2018deep,filos2019systematic,hendrycks2019benchmarking,ovadia2019can,dusenberry2020analyzing}. These all introduce a new task and evaluate a variety of baselines on that task. In practice, they are unmaintained, focusing on experimental insights rather than the codebase as the contribution. Our work provides an extensive set of benchmarks (in several cases, unifying the above ones), has a larger set of baselines across these benchmarks, and focuses on designing scalable, forkable, and well-tested code.

\vspace{-1ex}
\section{Uncertainty Baselines}

Uncertainty Baselines sets up each benchmark as a choice of base model, training dataset, and a suite of evaluation metrics.
\begin{enumerate}\setlength{\itemsep}{-0.5ex}
    \item 
Base models (architectures) include Wide ResNet 28-10 \citep{zagoruyko2016wide}, ResNet-50 \citep{he2016deep}, BERT \citep{devlin2018bert}, and simple MLPs.
    \item 
Training datasets include standard machine learning datasets -- CIFAR \citep{cifar10,cifar100}, ImageNet \citep{imagenet}, and UCI \citep{UCI} -- as well as more real-world problems -- Clinc Intent Detection \citep{clinc}, Kaggle's Diabetic Retinopathy Detection \citep{filos2019systematic}, and Wikipedia Toxicity \citep{wikipedia_talk}.
These span modalities such as tabular, text, and images.
    \item 
Evaluation includes predictive metrics such as accuracy, uncertainty metrics such as selective prediction and calibration error, compute metrics such as inference latency, and performance under in- and out-of-distribution datasets.
\end{enumerate}
As of this writing, we provide a total of 83 baselines, comprising 19 methods sweeping over standard and more recent strategies over 9 benchmarks.

\begin{figure}[!t]
\centering
\includegraphics[width=\textwidth]{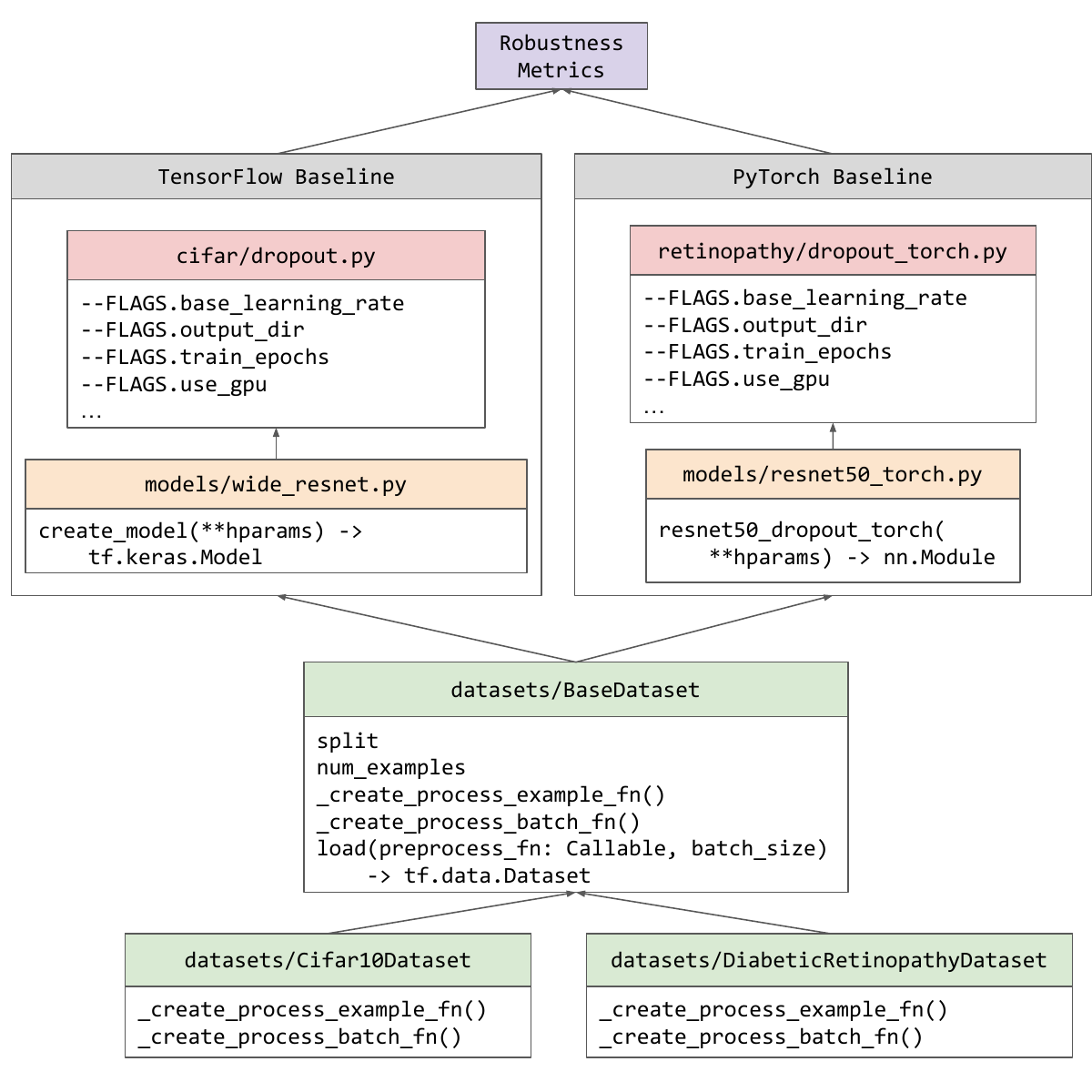}
\caption{The structure for an experiment under the TensorFlow or Pytorch backend. One instantiates a dataset (\texttt{Cifar10Dataset} or \texttt{DiabeticRetinopathyDataset}) and model (\texttt{wide\_resnet} or \texttt{resnet50\_torch}) within an end-to-end training script. After training, one inputs saved model checkpoints into Robustness Metrics for evaluation.}
\label{fig:uml}
\end{figure}

\textbf{Modularity.} In order to optimize for researchers to easily experiment on baselines (specifically, fork them), we designed the baselines to be as modular as possible and with minimal non-standard dependencies. API-wise, Uncertainty Baselines provides little to no abstractions: datasets are light wrappers around TensorFlow Datasets \citep{TFDS}, models are Keras models, and training/test logic is in raw TensorFlow \citep{tensorflow2015-whitepaper} This allows new users to more easily run individual examples, or incorporate our datasets and/or models into their libraries.
For out-of-distribution evaluation, we plug our trained models into Robustness Metrics \citep{djolonga2020robustness}.
\Cref{fig:uml} illustrates how the modules fit together.

\textbf{Framework.} Uncertainty Baselines is framework-agnostic.
The dataset and metric modules are NumPy-compatible, and interoperate in a performant manner with modern deep learning frameworks including TensorFlow, Jax, and PyTorch. For example, our baselines on the JFT-300M dataset use 
raw JAX, and we include a PyTorch Monte Carlo Dropout baseline on the Diabetic Retinopathy dataset. In practice, for ease of code and performance comparison, we choose a specific backend for each benchmark and develop all baselines under that backend (most often TensorFlow). Our Jax and PyTorch baselines demonstrate that implementation with other frameworks is supported and straightforward.

\begin{figure}[!tb]
\centering
 \begin{subfigure}{.49\textwidth}
  \centering
  \includegraphics[width=\textwidth]{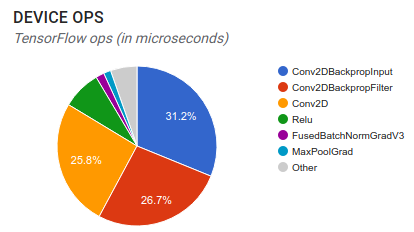}
\end{subfigure}%
\begin{subfigure}{.49\textwidth}
  \centering
  \includegraphics[width=\textwidth]{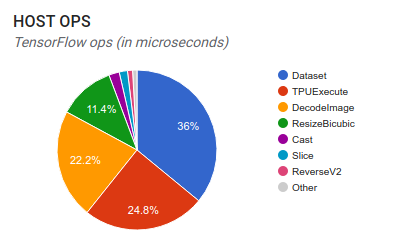}
\end{subfigure}
\vspace{-1.5ex}
\caption{Performance analysis of a MIMO baseline on a TPUv3-32 using the \href{https://www.tensorflow.org/tensorboard/tensorboard_profiling_keras}{TensorFlow Profiler}. The runtime is optimized, bound only by model operations, an irreducible bottleneck for a given baseline. Our implementations have 100\% utilization of the TPU devices.
\vspace{-2ex}
}
\label{fig:performance}
\end{figure}

\textbf{Hardware.} All baselines run on CPU, GPU, or Google Cloud TPUs.
Baselines are optimized for a default hardware configuration and often assume a memory requirement and number of chips (e.g., 1 GPU, or TPUv2-32) in order to reproduce the results. We employ the latest coding practices to fully utilize accelerator chips (\Cref{fig:performance}) so researchers can leverage the most performant baselines.

\textbf{Hyperparameters.} 
Hyperparameters and other experiment configuration values easily number in the dozens for a given baseline.
Uncertainty Baselines uses standard Python flags to specify hyperparameters, setting default values to reproduce best performance. Flags are simple, require no additional framework,  and
are easy to plug into other pipelines or extend.
We also document the protocol to properly tune and evaluate baselines---a common source of discrepancy in papers.

\textbf{Reproducibility.}
All modules include testing, and all results are reported over multiple seeds.
Computing metrics on trained models
can be prohibitively expensive let alone training from scratch. Therefore
we also provide TensorBoard dashboards which include all training, tuning, and evaluation metrics.
An example can be found \href{https://tensorboard.dev/experiment/1qy7JJfYQYqQ1lanieSYew/#scalars&runSelectionState=eyIxIjp0cnVlLCIyIjp0cnVlLCIzIjp0cnVlLCI0Ijp0cnVlLCI1Ijp0cnVlLCI2Ijp0cnVlLCI3Ijp0cnVlLCI4Ijp0cnVlLCI5Ijp0cnVlLCIxMCI6dHJ1ZSwiMTEiOnRydWUsIjEyIjp0cnVlLCIxMyI6dHJ1ZSwiMTQiOnRydWUsIjE1Ijp0cnVlLCIxNiI6dHJ1ZSwiMTciOnRydWUsIjE4Ijp0cnVlLCIxOSI6dHJ1ZSwiMjAiOnRydWUsIjIxIjp0cnVlLCIyMiI6dHJ1ZSwiMjMiOnRydWUsIjI0Ijp0cnVlLCIyNSI6dHJ1ZSwiMjYiOnRydWUsIjI3Ijp0cnVlLCIyOCI6dHJ1ZSwiMjkiOnRydWUsIjMwIjp0cnVlLCIzMSI6dHJ1ZSwiMzIiOnRydWUsIjMzIjp0cnVlLCIzNCI6dHJ1ZSwiMzUiOnRydWUsIjM2Ijp0cnVlLCIzNyI6dHJ1ZSwiMzgiOnRydWUsIjM5Ijp0cnVlLCI0MCI6dHJ1ZSwiNDEiOnRydWUsIjQyIjp0cnVlLCI0MyI6dHJ1ZSwiNDQiOnRydWUsIjQ1Ijp0cnVlLCI0NiI6dHJ1ZSwiNDciOnRydWUsIjQ4Ijp0cnVlLCI0OSI6dHJ1ZSwiNTAiOnRydWV9}{here}.

\begin{figure}[!tb]
\centering
\includegraphics[width=\linewidth]{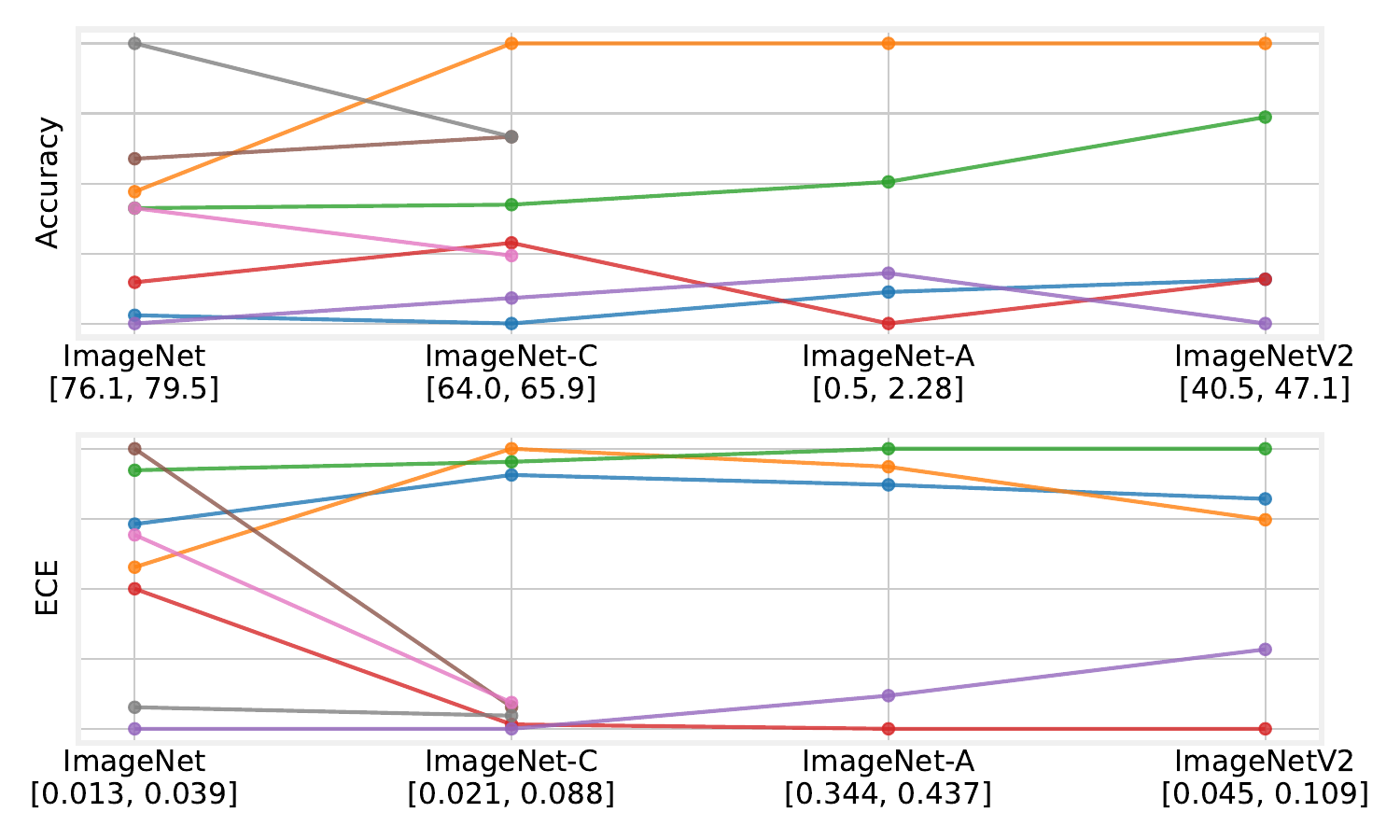}
\vspace{-4ex}
\caption{%
8 baselines evaluated on ImageNet, ImageNet-C, ImageNet-A, and ImageNetV2 (matched frequency variant).
\textbf{(top)} Top-1 accuracy.
\textbf{(bottom)} Expected calibration error.
Results demonstrate the many baselines available with competitive performance.
\vspace{-3ex}
}
\label{fig:imagenet}
\end{figure}

\begin{figure}[!tb]
\centering
\begin{subfigure}{.495\textwidth}
  \centering
  \includegraphics[width=\columnwidth]{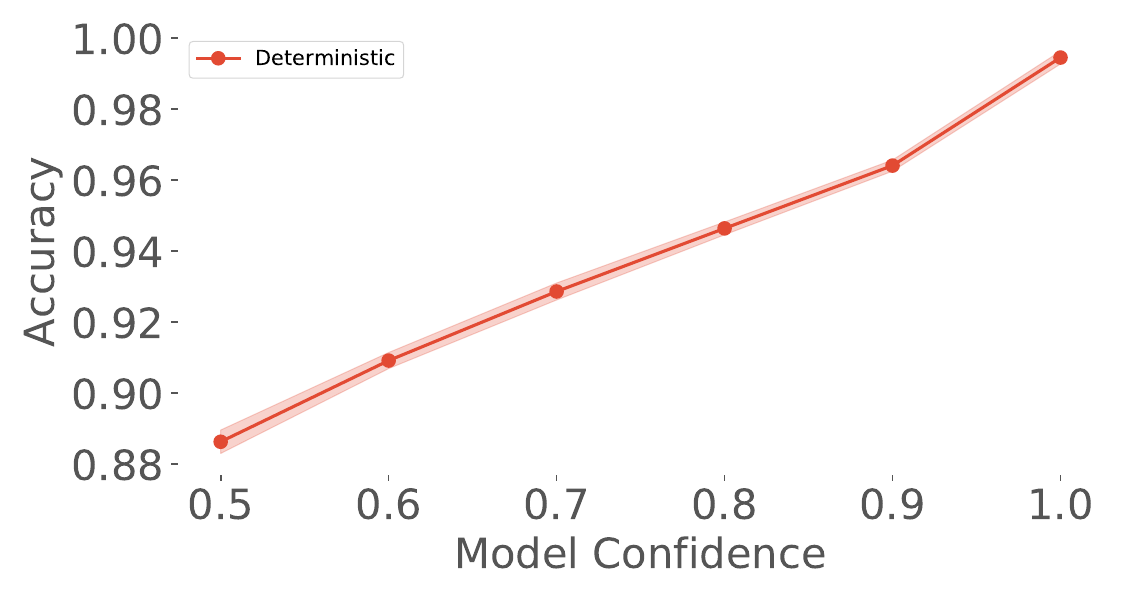}
\end{subfigure}%
\begin{subfigure}{.495\textwidth}
  \centering
  \includegraphics[width=\columnwidth]{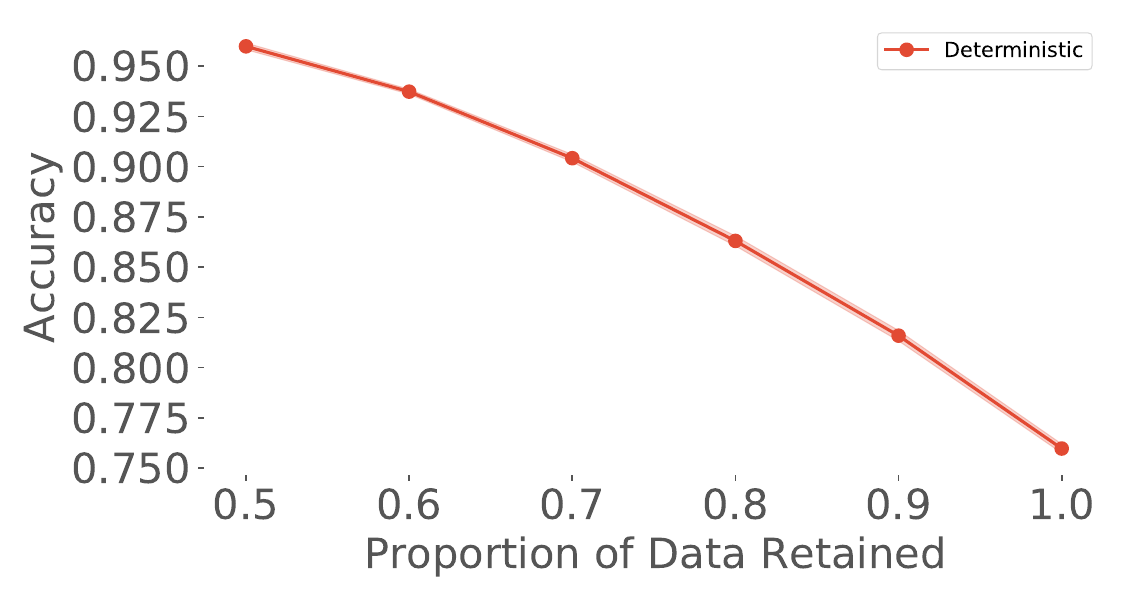}
\end{subfigure}
\vspace{-2ex}
\caption{ImageNet baseline applied to deferred prediction. In this task, one defers predictions according to the model's confidence (\textbf{left}) or a desired data retention rate (\textbf{right}).
}
\vspace{-2ex}
\label{fig:selective_prediction}
\end{figure}

\vspace{-2ex}
\section{Results}
To provide an example of Uncertainty Baselines' features, we display baselines available on 1 of 9 tasks: ImageNet. \Cref{fig:imagenet} displays accuracy and calibration error across 8 baselines, evaluated on in- and out-of-distribution.\footnote{We omit a legend to avoid drawing comparisons among which specific baselines perform best. See our full leaderboards to draw those insights
at \href{https://github.com/google/uncertainty-baselines/blob/master/baselines/imagenet/README.md}{\texttt{baselines/imagenet/README.md}}.
}
\Cref{fig:selective_prediction} provides an example of applying such baselines to a downstream task. Overall, the results demonstrate only a sampling of the repository's capabilities. We are excited to see new research already building on the baselines.

\newpage

\vskip 0.2in
\bibliography{bib}

\begin{thebibliography}{43}
\providecommand{\natexlab}[1]{#1}
\providecommand{\url}[1]{\texttt{#1}}
\expandafter\ifx\csname urlstyle\endcsname\relax
  \providecommand{\doi}[1]{doi: #1}\else
  \providecommand{\doi}{doi: \begingroup \urlstyle{rm}\Url}\fi

\bibitem[Abadi et~al.(2015)Abadi, Agarwal, Barham, Brevdo, Chen, Citro,
  Corrado, Davis, Dean, Devin, Ghemawat, Goodfellow, Harp, Irving, Isard, Jia,
  Jozefowicz, Kaiser, Kudlur, Levenberg, Man\'{e}, Monga, Moore, Murray, Olah,
  Schuster, Shlens, Steiner, Sutskever, Talwar, Tucker, Vanhoucke, Vasudevan,
  Vi\'{e}gas, Vinyals, Warden, Wattenberg, Wicke, Yu, and
  Zheng]{tensorflow2015-whitepaper}
Mart\'{\i}n Abadi, Ashish Agarwal, Paul Barham, Eugene Brevdo, Zhifeng Chen,
  Craig Citro, Greg~S. Corrado, Andy Davis, Jeffrey Dean, Matthieu Devin,
  Sanjay Ghemawat, Ian Goodfellow, Andrew Harp, Geoffrey Irving, Michael Isard,
  Yangqing Jia, Rafal Jozefowicz, Lukasz Kaiser, Manjunath Kudlur, Josh
  Levenberg, Dandelion Man\'{e}, Rajat Monga, Sherry Moore, Derek Murray, Chris
  Olah, Mike Schuster, Jonathon Shlens, Benoit Steiner, Ilya Sutskever, Kunal
  Talwar, Paul Tucker, Vincent Vanhoucke, Vijay Vasudevan, Fernanda Vi\'{e}gas,
  Oriol Vinyals, Pete Warden, Martin Wattenberg, Martin Wicke, Yuan Yu, and
  Xiaoqiang Zheng.
\newblock {TensorFlow}: Large-scale machine learning on heterogeneous systems,
  2015.
\newblock URL \url{https://www.tensorflow.org/}.
\newblock Software available from tensorflow.org.

\bibitem[Bello et~al.(2021)Bello, Fedus, Du, Cubuk, Srinivas, Lin, Shlens, and
  Zoph]{bello2021revisiting}
Irwan Bello, William Fedus, Xianzhi Du, Ekin~D Cubuk, Aravind Srinivas,
  Tsung-Yi Lin, Jonathon Shlens, and Barret Zoph.
\newblock Revisiting resnets: Improved training and scaling strategies.
\newblock \emph{arXiv preprint arXiv:2103.07579}, 2021.

\bibitem[Blundell et~al.(2015)Blundell, Cornebise, Kavukcuoglu, and
  Wierstra]{blundell2015weight}
Charles Blundell, Julien Cornebise, Koray Kavukcuoglu, and Daan Wierstra.
\newblock Weight uncertainty in neural network.
\newblock In \emph{International Conference on Machine Learning}, pages
  1613--1622. PMLR, 2015.

\bibitem[Carratino et~al.(2020)Carratino, Ciss{\'e}, Jenatton, and
  Vert]{carratino2020mixup}
Luigi Carratino, Moustapha Ciss{\'e}, Rodolphe Jenatton, and Jean-Philippe
  Vert.
\newblock On mixup regularization.
\newblock \emph{arXiv preprint arXiv:2006.06049}, 2020.

\bibitem[Collier et~al.(2021)Collier, Mustafa, Kokiopoulou, Jenatton, and
  Berent]{collier2020simple}
Mark Collier, Basil Mustafa, Efi Kokiopoulou, Rodolphe Jenatton, and Jesse
  Berent.
\newblock Correlated input-dependent label noise in large-scale image
  classification.
\newblock In \emph{Proceedings of the IEEE Conference on Computer Vision and
  Pattern Recognition}, 2021.

\bibitem[Conversation~AI(2017)]{toxiccomment}
Jigsaw Conversation~AI.
\newblock Toxic comment classification challenge.
\newblock
  \url{https://www.kaggle.com/c/jigsaw-toxic-comment-classification-challenge},
  2017.

\bibitem[D'Amour et~al.(2020)D'Amour, Heller, Moldovan, Adlam, Alipanahi,
  Beutel, Chen, Deaton, Eisenstein, Hoffman, et~al.]{d2020underspecification}
Alexander D'Amour, Katherine Heller, Dan Moldovan, Ben Adlam, Babak Alipanahi,
  Alex Beutel, Christina Chen, Jonathan Deaton, Jacob Eisenstein, Matthew~D
  Hoffman, et~al.
\newblock Underspecification presents challenges for credibility in modern
  machine learning.
\newblock \emph{arXiv preprint arXiv:2011.03395}, 2020.

\bibitem[Devlin et~al.(2018)Devlin, Chang, Lee, and Toutanova]{devlin2018bert}
Jacob Devlin, Ming-Wei Chang, Kenton Lee, and Kristina Toutanova.
\newblock Bert: Pre-training of deep bidirectional transformers for language
  understanding.
\newblock \emph{arXiv preprint arXiv:1810.04805}, 2018.

\bibitem[Dhariwal et~al.(2017)Dhariwal, Hesse, Klimov, Nichol, Plappert,
  Radford, Schulman, Sidor, Wu, and Zhokhov]{dhariwal2017openai}
Prafulla Dhariwal, Christopher Hesse, Oleg Klimov, Alex Nichol, Matthias
  Plappert, Alec Radford, John Schulman, Szymon Sidor, Yuhuai Wu, and Peter
  Zhokhov.
\newblock Openai baselines.
\newblock \url{https://github.com/openai/baselines}, 2017.

\bibitem[Djolonga et~al.(2020)Djolonga, Hubis, Minderer, Nado, Nixon,
  Romijnders, Tran, and Lucic]{djolonga2020robustness}
Josip Djolonga, Frances Hubis, Matthias Minderer, Zack Nado, Jeremy Nixon, Rob
  Romijnders, Dustin Tran, and Mario Lucic.
\newblock {R}obustness {M}etrics, 2020.
\newblock URL \url{https://github.com/google-research/robustness_metrics}.

\bibitem[Dua and Graff(2017)]{UCI}
Dheeru Dua and Casey Graff.
\newblock {UCI} machine learning repository, 2017.
\newblock URL \url{http://archive.ics.uci.edu/ml}.

\bibitem[Dusenberry et~al.(2020{\natexlab{a}})Dusenberry, Jerfel, Wen, Ma,
  Snoek, Heller, Lakshminarayanan, and Tran]{dusenberry2020efficient}
Michael Dusenberry, Ghassen Jerfel, Yeming Wen, Yian Ma, Jasper Snoek,
  Katherine Heller, Balaji Lakshminarayanan, and Dustin Tran.
\newblock Efficient and scalable bayesian neural nets with rank-1 factors.
\newblock In \emph{International conference on machine learning}, pages
  2782--2792. PMLR, 2020{\natexlab{a}}.

\bibitem[Dusenberry et~al.(2020{\natexlab{b}})Dusenberry, Tran, Choi, Kemp,
  Nixon, Jerfel, Heller, and Dai]{dusenberry2020analyzing}
Michael~W Dusenberry, Dustin Tran, Edward Choi, Jonas Kemp, Jeremy Nixon,
  Ghassen Jerfel, Katherine Heller, and Andrew~M Dai.
\newblock Analyzing the role of model uncertainty for electronic health
  records.
\newblock In \emph{Proceedings of the ACM Conference on Health, Inference, and
  Learning}, pages 204--213, 2020{\natexlab{b}}.

\bibitem[Farquhar et~al.(2020)Farquhar, Osborne, and Gal]{farquhar2020radial}
Sebastian Farquhar, Michael~A Osborne, and Yarin Gal.
\newblock Radial bayesian neural networks: Beyond discrete support in
  large-scale bayesian deep learning.
\newblock In \emph{International Conference on Artificial Intelligence and
  Statistics}, pages 1352--1362. PMLR, 2020.

\bibitem[Filos et~al.(2019)Filos, Farquhar, Gomez, Rudner, Kenton, Smith,
  Alizadeh, de~Kroon, and Gal]{filos2019systematic}
Angelos Filos, Sebastian Farquhar, Aidan~N Gomez, Tim~GJ Rudner, Zachary
  Kenton, Lewis Smith, Milad Alizadeh, Arnoud de~Kroon, and Yarin Gal.
\newblock A systematic comparison of bayesian deep learning robustness in
  diabetic retinopathy tasks.
\newblock \emph{arXiv preprint arXiv:1912.10481}, 2019.

\bibitem[Gal and Ghahramani(2016)]{gal2016dropout}
Yarin Gal and Zoubin Ghahramani.
\newblock Dropout as a bayesian approximation: Representing model uncertainty
  in deep learning.
\newblock In \emph{international conference on machine learning}, pages
  1050--1059. PMLR, 2016.

\bibitem[Havasi et~al.(2020)Havasi, Jenatton, Fort, Liu, Snoek,
  Lakshminarayanan, Dai, and Tran]{havasi2020training}
Marton Havasi, Rodolphe Jenatton, Stanislav Fort, Jeremiah~Zhe Liu, Jasper
  Snoek, Balaji Lakshminarayanan, Andrew~M Dai, and Dustin Tran.
\newblock Training independent subnetworks for robust prediction.
\newblock \emph{arXiv preprint arXiv:2010.06610}, 2020.

\bibitem[He et~al.(2016)He, Zhang, Ren, and Sun]{he2016deep}
Kaiming He, Xiangyu Zhang, Shaoqing Ren, and Jian Sun.
\newblock Deep residual learning for image recognition.
\newblock In \emph{Proceedings of the IEEE conference on computer vision and
  pattern recognition}, pages 770--778, 2016.

\bibitem[Hendrycks and Dietterich(2019)]{hendrycks2019benchmarking}
Dan Hendrycks and Thomas Dietterich.
\newblock Benchmarking neural network robustness to common corruptions and
  perturbations.
\newblock \emph{arXiv preprint arXiv:1903.12261}, 2019.

\bibitem[Krizhevsky(2009)]{cifar}
Alex Krizhevsky.
\newblock Learning multiple layers of features from tiny images.
\newblock Technical report, University of Toronto, 2009.

\bibitem[Krizhevsky et~al.({\natexlab{a}})Krizhevsky, Nair, and
  Hinton]{cifar10}
Alex Krizhevsky, Vinod Nair, and Geoffrey Hinton.
\newblock Cifar-10 (canadian institute for advanced research).
\newblock {\natexlab{a}}.
\newblock URL \url{http://www.cs.toronto.edu/~kriz/cifar.html}.

\bibitem[Krizhevsky et~al.({\natexlab{b}})Krizhevsky, Nair, and
  Hinton]{cifar100}
Alex Krizhevsky, Vinod Nair, and Geoffrey Hinton.
\newblock Cifar-100 (canadian institute for advanced research).
\newblock {\natexlab{b}}.
\newblock URL \url{http://www.cs.toronto.edu/~kriz/cifar.html}.

\bibitem[Kurach et~al.(2019)Kurach, Lu{\v{c}}i{\'c}, Zhai, Michalski, and
  Gelly]{kurach2019large}
Karol Kurach, Mario Lu{\v{c}}i{\'c}, Xiaohua Zhai, Marcin Michalski, and
  Sylvain Gelly.
\newblock A large-scale study on regularization and normalization in gans.
\newblock In \emph{International Conference on Machine Learning}, pages
  3581--3590. PMLR, 2019.

\bibitem[Lakshminarayanan et~al.(2016)Lakshminarayanan, Pritzel, and
  Blundell]{lakshminarayanan2016simple}
Balaji Lakshminarayanan, Alexander Pritzel, and Charles Blundell.
\newblock Simple and scalable predictive uncertainty estimation using deep
  ensembles.
\newblock \emph{arXiv preprint arXiv:1612.01474}, 2016.

\bibitem[Larson et~al.(2019)Larson, Mahendran, Peper, Clarke, Lee, Hill,
  Kummerfeld, Leach, Laurenzano, Tang, et~al.]{clinc}
Stefan Larson, Anish Mahendran, Joseph~J Peper, Christopher Clarke, Andrew Lee,
  Parker Hill, Jonathan~K Kummerfeld, Kevin Leach, Michael~A Laurenzano,
  Lingjia Tang, et~al.
\newblock An evaluation dataset for intent classification and out-of-scope
  prediction.
\newblock \emph{arXiv preprint arXiv:1909.02027}, 2019.

\bibitem[LeCun and Cortes(2010)]{mnist}
Yann LeCun and Corinna Cortes.
\newblock {MNIST} handwritten digit database.
\newblock 2010.
\newblock URL \url{http://yann.lecun.com/exdb/mnist/}.

\bibitem[Liu et~al.(2020)Liu, Lin, Padhy, Tran, Bedrax-Weiss, and
  Lakshminarayanan]{liu2020simple}
Jeremiah~Zhe Liu, Zi~Lin, Shreyas Padhy, Dustin Tran, Tania Bedrax-Weiss, and
  Balaji Lakshminarayanan.
\newblock Simple and principled uncertainty estimation with deterministic deep
  learning via distance awareness.
\newblock \emph{arXiv preprint arXiv:2006.10108}, 2020.

\bibitem[Loshchilov and Hutter(2017)]{loshchilov2017decoupled}
Ilya Loshchilov and Frank Hutter.
\newblock Decoupled weight decay regularization.
\newblock \emph{arXiv preprint arXiv:1711.05101}, 2017.

\bibitem[Melis et~al.(2017)Melis, Dyer, and Blunsom]{melis2017state}
G{\'a}bor Melis, Chris Dyer, and Phil Blunsom.
\newblock On the state of the art of evaluation in neural language models.
\newblock \emph{arXiv preprint arXiv:1707.05589}, 2017.

\bibitem[Mukhoti et~al.(2019)Mukhoti, Kulharia, Sanyal, Golodetz, Torr, and
  Dokania]{mukhoti2019intriguing}
Jishnu Mukhoti, Viveka Kulharia, Amartya Sanyal, Stuart Golodetz, Philip Torr,
  and Puneet Dokania.
\newblock The intriguing effects of focal loss on the calibration of deep
  neural networks.
\newblock 2019.

\bibitem[Nado et~al.(2021)Nado, Gilmer, Shallue, Anil, and Dahl]{nado2021large}
Zachary Nado, Justin~M Gilmer, Christopher~J Shallue, Rohan Anil, and George~E
  Dahl.
\newblock A large batch optimizer reality check: Traditional, generic
  optimizers suffice across batch sizes.
\newblock \emph{arXiv preprint arXiv:2102.06356}, 2021.

\bibitem[Ovadia et~al.(2019)Ovadia, Fertig, Ren, Nado, Sculley, Nowozin,
  Dillon, Lakshminarayanan, and Snoek]{ovadia2019can}
Yaniv Ovadia, Emily Fertig, Jie Ren, Zachary Nado, David Sculley, Sebastian
  Nowozin, Joshua~V Dillon, Balaji Lakshminarayanan, and Jasper Snoek.
\newblock Can you trust your model's uncertainty? evaluating predictive
  uncertainty under dataset shift.
\newblock \emph{arXiv preprint arXiv:1906.02530}, 2019.

\bibitem[Riquelme et~al.(2018)Riquelme, Tucker, and Snoek]{riquelme2018deep}
Carlos Riquelme, George Tucker, and Jasper Snoek.
\newblock Deep bayesian bandits showdown: An empirical comparison of bayesian
  deep networks for thompson sampling.
\newblock \emph{arXiv preprint arXiv:1802.09127}, 2018.

\bibitem[Russakovsky et~al.(2015)Russakovsky, Deng, Su, Krause, Satheesh, Ma,
  Huang, Karpathy, Khosla, Bernstein, Berg, and Fei-Fei]{imagenet}
Olga Russakovsky, Jia Deng, Hao Su, Jonathan Krause, Sanjeev Satheesh, Sean Ma,
  Zhiheng Huang, Andrej Karpathy, Aditya Khosla, Michael Bernstein,
  Alexander~C. Berg, and Li~Fei-Fei.
\newblock {ImageNet Large Scale Visual Recognition Challenge}.
\newblock \emph{International Journal of Computer Vision (IJCV)}, 115\penalty0
  (3):\penalty0 211--252, 2015.

\bibitem[Sinha et~al.(2020)Sinha, Pineau, Forde, Ke, and
  Larochelle]{sinha2020neurips}
Koustuv Sinha, Joelle Pineau, Jessica Forde, Rosemary~Nan Ke, and Hugo
  Larochelle.
\newblock Neurips 2019 reproducibility challenge.
\newblock \emph{ReScience C}, 6\penalty0 (2):\penalty0 11, 2020.

\bibitem[Sutskever et~al.(2013)Sutskever, Martens, Dahl, and
  Hinton]{sutskever2013importance}
Ilya Sutskever, James Martens, George Dahl, and Geoffrey Hinton.
\newblock On the importance of initialization and momentum in deep learning.
\newblock In \emph{International conference on machine learning}, pages
  1139--1147. PMLR, 2013.

\bibitem[Szegedy et~al.(2015)Szegedy, Liu, Jia, Sermanet, Reed, Anguelov,
  Erhan, Vanhoucke, and Rabinovich]{szegedy2015going}
Christian Szegedy, Wei Liu, Yangqing Jia, Pierre Sermanet, Scott Reed, Dragomir
  Anguelov, Dumitru Erhan, Vincent Vanhoucke, and Andrew Rabinovich.
\newblock Going deeper with convolutions.
\newblock In \emph{Proceedings of the IEEE conference on computer vision and
  pattern recognition}, pages 1--9, 2015.

\bibitem[Tan and Le(2019)]{tan2019efficientnet}
Mingxing Tan and Quoc Le.
\newblock Efficientnet: Rethinking model scaling for convolutional neural
  networks.
\newblock In \emph{International Conference on Machine Learning}, pages
  6105--6114. PMLR, 2019.

\bibitem[{TFDS Team}()]{TFDS}
{TFDS Team}.
\newblock {TensorFlow Datasets}, a collection of ready-to-use datasets.
\newblock \url{https://www.tensorflow.org/datasets}.

\bibitem[Wen et~al.(2020)Wen, Tran, and Ba]{wen2020batchensemble}
Yeming Wen, Dustin Tran, and Jimmy Ba.
\newblock Batchensemble: an alternative approach to efficient ensemble and
  lifelong learning.
\newblock \emph{arXiv preprint arXiv:2002.06715}, 2020.

\bibitem[Wenzel et~al.(2020)Wenzel, Snoek, Tran, and
  Jenatton]{wenzel2020hyperparameter}
Florian Wenzel, Jasper Snoek, Dustin Tran, and Rodolphe Jenatton.
\newblock Hyperparameter ensembles for robustness and uncertainty
  quantification.
\newblock \emph{arXiv preprint arXiv:2006.13570}, 2020.

\bibitem[Wulczyn et~al.(2017)Wulczyn, Thain, and Dixon]{wikipedia_talk}
Ellery Wulczyn, Nithum Thain, and Lucas Dixon.
\newblock Ex machina: Personal attacks seen at scale.
\newblock In \emph{Proceedings of the 26th International Conference on World
  Wide Web}, WWW '17, pages 1391--1399, Republic and Canton of Geneva, CHE,
  2017. International World Wide Web Conferences Steering Committee.
\newblock ISBN 9781450349130.
\newblock \doi{10.1145/3038912.3052591}.
\newblock URL \url{https://doi.org/10.1145/3038912.3052591}.

\bibitem[Zagoruyko and Komodakis(2016)]{zagoruyko2016wide}
Sergey Zagoruyko and Nikos Komodakis.
\newblock Wide residual networks.
\newblock \emph{arXiv preprint arXiv:1605.07146}, 2016.

\end{thebibliography}

\appendix

\newpage

\section{Dataset Details}
For CIFAR10 and CIFAR100, we padded the images with 4 pixels of 0's before doing a random crop to 32x32 pixels, followed by a left-right flip with 50\% chance. For ImageNet, we used ResNet preprocessing as described in \citet{he2016deep}, but also support the common Inception preprocessing from \citet{szegedy2015going}. All preprocessing is deterministic given a random seed, using \texttt{tf.random.experimental.stateless\_split} and \texttt{tf.random.experimental.stateless\_fold\_in}. For the Diabetic Retinopathy benchmarks we used the Kaggle competition dataset as in \citet{filos2019systematic}.

\section{Model Details}
For CIFAR10 and CIFAR100 we provide methods based on the Wide ResNet models, typically the Wide ResNet-28 size \citep{zagoruyko2016wide}.
For ImageNet and the Diabetic Retinopathy benchmarks, we provide methods based on the ResNet-50 model \citep{he2016deep}. For ImageNet we additionally use methods based on the EfficientNet models \citep{tan2019efficientnet}. For the Toxic Comments and CLINC Intent Detection benchmarks, our methods are based on the BERT-Base model \citep{devlin2018bert}.

\section{Hyperparameter Tuning}\label{appendix:hparams}
All image benchmarks were trained with Nesterov momentum \citep{sutskever2013importance}, except for the EfficientNet models which use RMSProp with $\rho = 0.9, \epsilon=\num{e-3}$. The text benchmarks were trained with the AdamW optimizer \citep{loshchilov2017decoupled} with a $\beta_2 = 0.999, \epsilon = \num{e-6}$. Unless otherwise noted, the image benchmarks used a linear warmup followed by a stepwise decay schedule, except for the EfficientNet models which used a linear warmup followed by an exponential decay. The text benchmarks used a linear warmup followed by a linear decay.

For the CIFAR10, CIFAR100, ImageNet, Toxic Comments, and CLINC Intent Detection benchmarks, the papers for each method contain their tuning details.

\paragraph{Diabetic Retinopathy benchmark tuning details.} For the Diabetic Retinopathy benchmark, we also provide our tuning results so that others can more easily retune their own methods. We conducted two rounds of quasirandom search on several hyperparameters (learning rate, momentum, dropout, variational posteriors, L2 regularization), where the first round was a heuristically-picked larger search space and the second round was a hand-tuned smaller range around the better performing values. Each round was for 50 trials, and the final hyperparameters were selected using the final validation AUC from the second tuning round. We finally retrained this best hyperparameter setting on the combined train and validation sets.

\newcommand{\specialcell}[2][c]{%
  \begin{tabular}[#1]{@{}c@{}}#2\end{tabular}}
  
\section{Supported Baselines}

\newpage

\begin{table}[!t]
\centering
\begin{tabular}{|c|c|}
\hline
Dataset & Method \\ \hline
\href{https://github.com/google/uncertainty-baselines/tree/master/baselines/cifar}{CIFAR} \citep{cifar} & BatchEnsemble \citep{wen2020batchensemble} \\ \hline
& Hyper-BatchEnsemble \citep{wenzel2020hyperparameter} \\ \hline
& MIMO \citep{havasi2020training} \\ \hline
& \specialcell{Rank-1 BNN (Gaussian)\\\citep{dusenberry2020efficient}} \\ \hline
& Rank-1 BNN (Cauchy) \\ \hline
& SNGP \citep{liu2020simple} \\ \hline
& \specialcell{MC-Dropout\\\citep{gal2016dropout}} \\ \hline
& \specialcell{Ensemble\\\citep{lakshminarayanan2016simple}} \\ \hline
& Hyper-deep ensemble \citep{wenzel2020hyperparameter} \\ \hline
& Variational Inference \citep{blundell2015weight} \\ \hline
& Heteroscedastic \citep{collier2020simple} \\ \hline
\hline
\href{https://github.com/google/uncertainty-baselines/tree/master/baselines/clinc_intent}{CLINC} \citep{clinc} & SNGP \\ \hline
& MC-Dropout \\ \hline
& Ensemble \\ \hline
\hline
\href{https://github.com/google/uncertainty-baselines/tree/master/baselines/diabetic_retinopathy_detection}{Diabetic Retinopathy Detection} \citep{filos2019systematic} & MC-Dropout \\ \hline
& Ensemble \\ \hline
& Radial BNNs \citep{farquhar2020radial} \\ \hline
& Variational Inference \\ \hline
\hline
\href{https://github.com/google/uncertainty-baselines/tree/master/baselines/imagenet}{ImageNet} \citep{imagenet} & MixUp \citep{carratino2020mixup} \\ \hline
& BatchEnsemble \\ \hline
& Hyper-BatchEnsemble \\ \hline
& MIMO \\ \hline
& Rank-1 BNN (Gaussian) \\ \hline
& Rank-1 BNN (Cauchy) \\ \hline
& SNGP \\ \hline
& MC-Dropout \\ \hline
& Ensemble \\ \hline
& Hyper-deep ensemble \\ \hline
& Variational Inference \\ \hline
& Heteroscedastic \\ \hline
\hline
\href{https://github.com/google/uncertainty-baselines/tree/master/baselines/mnist}{MNIST} \citep{mnist} & Variational Inference \\ \hline
\hline
\href{https://github.com/google/uncertainty-baselines/tree/master/baselines/mnist}{Toxic Comments Detection} \citep{toxiccomment} & SNGP \\ \hline
& MC-Dropout \\ \hline
& Ensemble \\ \hline
& Focal Loss \citep{mukhoti2019intriguing} \\ \hline
\hline
\href{https://github.com/google/uncertainty-baselines/tree/master/baselines/uci}{UCI} \citep{UCI} & Variational Inference \\ \hline
\end{tabular}
\caption{Currently implemented methods for each dataset, in addition to a deterministic baseline. See repository for a more updated list.}
\label{table:all-baselines}
\end{table}

\FloatBarrier

\section{Open-Source Data}\label{appendix:tensorboard}
The tuning and final metrics data for the Diabetic Retinopathy benchmarks can be found at the following URLs:
\begin{itemize}
\item \href{https://tensorboard.dev/experiment/nAygVvdjSWWAEQRDD8Z0Aw/}{Deterministic First Tuning}
\item \href{https://tensorboard.dev/experiment/GLxGQR8pQhypBr9jGdBMUQ/}{Deterministic Final Tuning}
\item \href{https://tensorboard.dev/experiment/lh5yXcwzRc2ZNmId34ujPw/}{Deterministic 10 seeds}
\item \href{https://tensorboard.dev/experiment/xDVLkDAgR1uJqyxIqkdPIQ/}{Dropout First Tuning}
\item \href{https://tensorboard.dev/experiment/1qy7JJfYQYqQ1lanieSYew/}{Dropout Final Tuning}
\item \href{https://tensorboard.dev/experiment/aMr4glcES6qg43P4HvckTg/}{Dropout 10 seeds}
\item \href{https://tensorboard.dev/experiment/gVwRJIRoQoyRrfG1boJVPA/}{Variational Inference First Tuning}
\item \href{https://tensorboard.dev/experiment/n9NYA7ryRG6jCYdpyQYoOQ/}{Variational Inference Final Tuning}
\item \href{https://tensorboard.dev/experiment/mPZt9k0lQ1yF2TAuE2cxqw/}{Variational Inference 10 seeds}
\item \href{https://tensorboard.dev/experiment/5CzJYikVTvKQLdqSnmUrpg/}{Radial BNN First Tuning}
\item \href{https://tensorboard.dev/experiment/RDf1PKZkSZ2PGo1H8wnWBw/}{Radial BNN Final Tuning}
\item \href{https://tensorboard.dev/experiment/040rBdKBQPir8cDhReyk3A/}{Radial BNN 10 seeds}
\end{itemize}

\end{document}